\documentclass{article}
\usepackage{spconf,graphicx,hyperref}
\usepackage{enumitem}
\usepackage{amsfonts}
\usepackage{multirow}
\usepackage{booktabs}
\usepackage{makecell}
\usepackage{caption}  
\usepackage{cite}
\usepackage{amsmath}

\title{FTIN: Frequency-Time Integration Network for Inertial Odometry}
\name{Shanshan Zhang$^{\star \dagger}$ \qquad Qi Zhang$^{\star}$ \qquad Siyue Wang$^{\star}$ \qquad Liqin Wu$^{\star}$  \qquad Tianshui Wen$^{\star}$  Ziheng Zhou$^{\star}$ 
\qquad Ao Peng$^{\star}$ \qquad Xuemin Hong$^{\star}$ \qquad Lingxiang Zheng$^{\star \ddagger}$ \qquad Yu Yang$^{\dagger \ddagger}$}
\address{$^{\star}$ Department of Information and Communication Engineering, Xiamen University, China \\
         $^{\dagger}$ Department of Electronic Science, Xiamen University, China\\
         $^{\ddagger}$ Corresponding author, lxzheng@xmu.edu.cn, yuyang15@xmu.edu.cn}

\begin{document}
\maketitle

\begin{abstract}
Inertial odometry (IO) leverages inertial measurement unit (IMU) signals for cost-effective localization. However, high IMU sampling rates introduce substantial redundancy that impedes IO's ability to attend to salient components, thereby creating an information bottleneck. To address this challenge, we propose a cross-domain IO framework that fuses information from the frequency and time domains. Specifically, we exploit the global context and energy-compaction properties of frequency-domain representations to capture holistic motion patterns and alleviate the bottleneck. To the best of our knowledge, this is among the first attempts to incorporate frequency-domain feature processing into IO. Experimental results on multiple public datasets demonstrate the effectiveness of the proposed frequency--time-domain fusion strategy.
\end{abstract}


\section{Introduction}
Inertial odometry (IO) aims to reconstruct motion trajectories from high-frequency inertial measurement unit (IMU) signals—comprising tri-axial accelerometer and gyroscope data—in order to enable low-cost and robust localization\cite{Tartan-IMU,AirIO}. IO has been widely adopted in applications such as autonomous driving and localization tasks\cite{InertialNavigationMeetsDeepLearning,InertialNet}.

Conventional IO methods estimate position and orientation based on Newtonian mechanics\cite{surveyILS}. However, the accumulation of errors due to sensor noise leads to substantial drift over time\cite{SINS}. Although incorporating physical priors can mitigate this drift, such approaches often restrict the method's applicability to specific environments\cite{PDR,CHE}. 

In recent years, the integration of machine learning techniques has improved the accuracy of IO systems and broadened their deployment scenarios\cite{RIDI,IDOL,Oxiod}.
A common strategy involves using CNNs to extract local motion features from IMU signals\cite{RoNIN}. However, CNN-based approaches tend to neglect long-range temporal dependencies due to their inherently limited receptive field, which compromises performance in dynamic contexts\cite{iMOT}. To address this limitation, CNNs have been combined with LSTM networks or attention mechanisms to capture contextual motion information more effectively\cite{RNIN-VIO,SCHNN,SSHNN,CTIN,iMOT}.
Despite these advancements, existing methods frequently overlook the information bottleneck posed by high-sampling-rate IMU signals. This bottleneck impairs the network’s ability to emphasize informative signal components\cite{FreTS}, thereby degrading the accuracy of trajectory reconstruction.

\begin{figure}[t]
\centering
\captionsetup{aboveskip=2pt,font=small}
\includegraphics[width=0.44\textwidth]{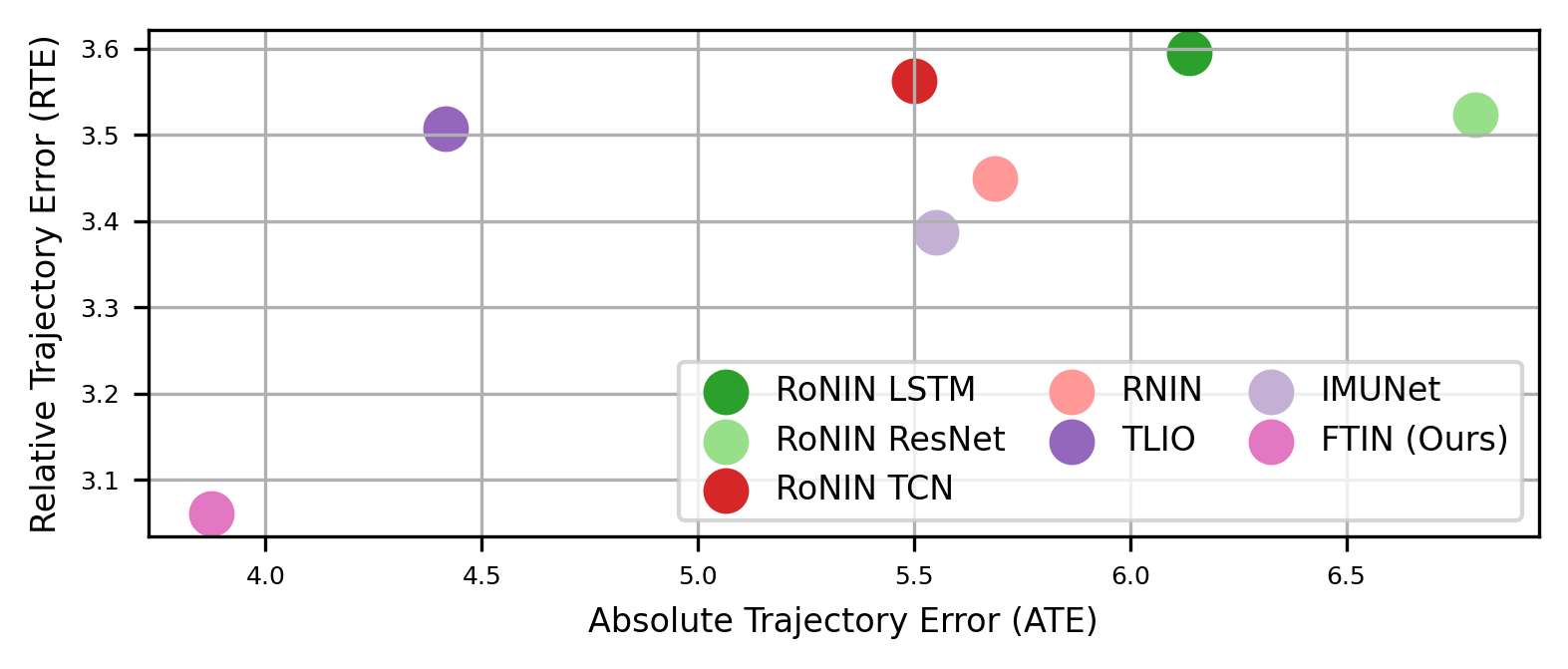}
\caption{Comparison on the RoNIN dataset. Algorithms closer to the lower-left corner indicate lower errors.}
\label{demo}
\vspace{-22pt}
\end{figure}

Beyond IO, both theoretical analyses and empirical studies in time-series modeling have demonstrated that frequency-domain representations offer improved global context awareness~\cite{FreTS}. By highlighting a small number of dominant frequency components, these representations reduce redundancy and enhance representational efficiency.
Motivated by these insights, we propose transforming IMU signals into the frequency domain to better capture contextual motion information. This transformation leverages the global temporal context inherent in the frequency domain and emphasizes dominant frequency components, making critical motion patterns more salient and facilitating effective feature extraction. The effectiveness of the proposed method is demonstrated in Fig.~\ref{demo}.
The main contributions of this work are as follows:
\begin{itemize}[noitemsep, nolistsep, leftmargin=*]
  \item Reinterpret IMU signals from a frequency-domain perspective to reduce redundancy and enable contextual motion modeling;
  \item Enhance the ResNet architecture with frequency-domain learning to integrate local motion features and global contextual dependencies;
  \item Introduce a Scalar LSTM module to improve time-domain contextual motion modeling and facilitate effective cross-domain feature fusion;
\end{itemize}

\begin{figure}[t] 
\centering
\captionsetup{aboveskip=2pt,font=small} 
\includegraphics[width=0.48\textwidth]{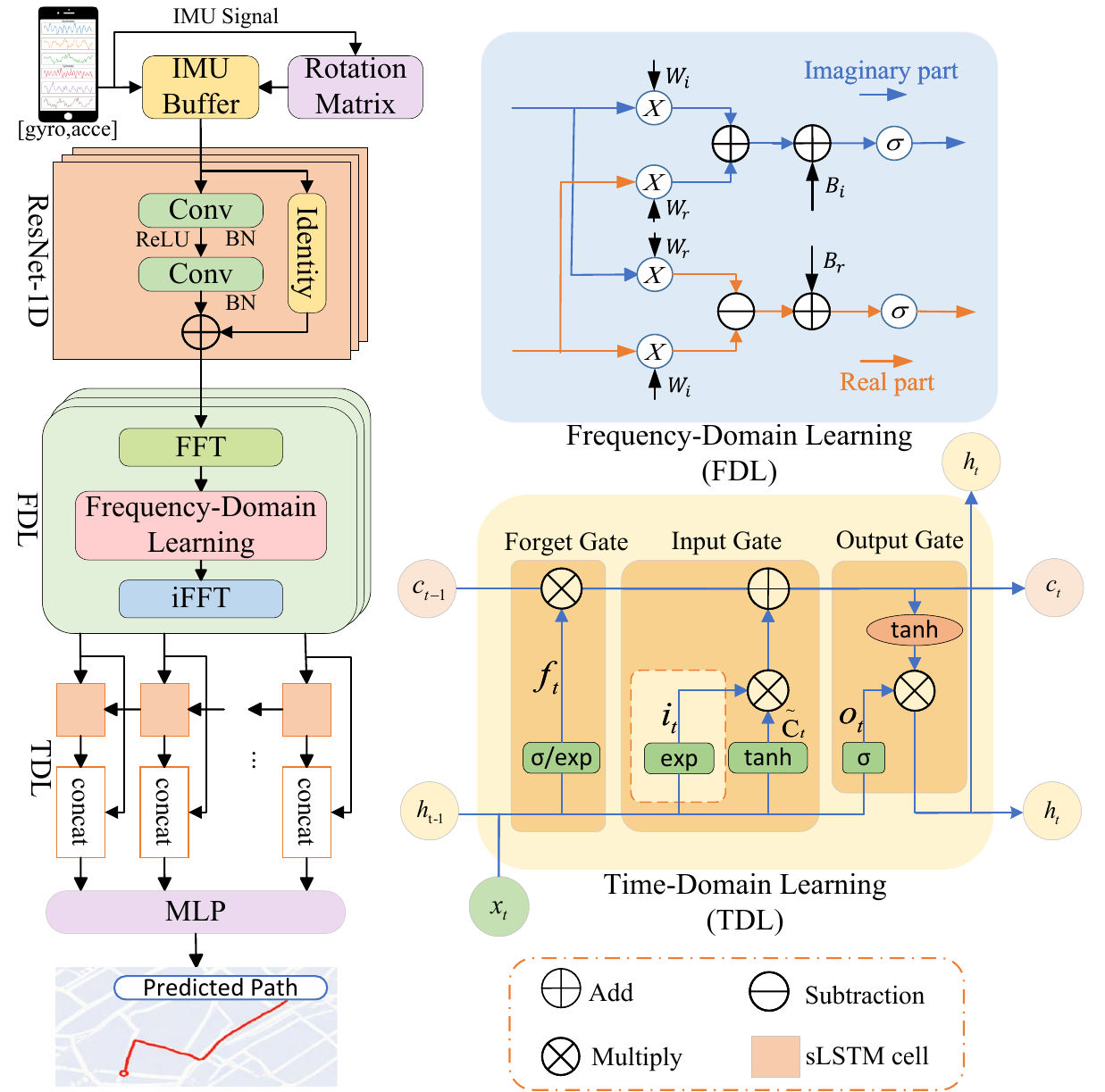}
\caption{Network Architecture Diagram of FTIN.}
\label{Schematic_Diagram}
\vspace{-20pt} 
\end{figure}

\section{Proposed Method}
\subsection{Overall Structure}
An overview of the proposed Frequency-Time Integration Network (FTIN) pipeline is illustrated in Fig.~\ref{Schematic_Diagram}. 

FTIN processes IMU signals through three successive stages to extract latent motion features. Prior research has demonstrated that shallow network layers primarily capture high-frequency local patterns, whereas deeper layers focus on low-frequency, long-range dependencies~\cite{InceptionTransformer}.
Based on this insight, we first apply a ResNet-1D to extract fine-grained motion features and project the signals into a higher-dimensional space.

In the second stage, a Fast Fourier transform (FFT) is applied to convert the features into the frequency domain. A multilayer perceptron (MLP) then learns global motion representations based on the transformed signals. This design leverages the global temporal context inherent in frequency-domain representations and the concentration of signal energy in a limited number of dominant frequency components. Frequency-Domain Learning is independently applied along both the channel and temporal dimensions to extract multidimensional global motion features.

In the third stage, we employ a Scalar LSTM (sLSTM)~\cite{beck:24xlstm} to capture global motion representations in the time domain. The scalar update mechanism of sLSTM preserves critical temporal information over long sequences, enhancing the model’s ability to learn sequential motion patterns.

Finally, the learned feature representations are mapped to predicted velocity values using an MLP. The entire network is trained by minimizing the mean squared error (MSE) between the predicted and ground-truth velocities.

\subsection{Frequency-Domain Learning}

Frequency-domain learning (FDL) offers a global perspective for capturing comprehensive motion representations and enables energy compaction, thereby alleviating the information bottleneck problem~\cite{FreTS}.

Given IMU signals within a time window, denoted as \( \mathbf{X} \in \mathbb{R}^{C \times L} \), where \( C = 6 \) represents tri-axial accelerometer and gyroscope signals (i.e., six channels) and \( L \) is the window length (numerically equal to the IMU sampling rate), the output of the ResNet-1D encoder is represented as \( \mathbf{X}_{\text{Res}} \in \mathbb{R}^{C_{\text{Res}} \times L_{\text{Res}}} \). To facilitate FDL, an additional embedding dimension \( d \) is introduced:
\begin{equation}
\mathbf{X}_{\text{tokenEmb}} = \text{reshape}(\mathbf{X}_{\text{Res}}) \times \mathbf{W}_1 \in \mathbb{R}^{C_{\text{Res}} \times L_{\text{Res}} \times d}
\end{equation}

Here, \( \text{reshape}(\cdot) \) introduces the embedding dimension without altering the original signal values, and \( \mathbf{W}_1 \in \mathbb{R}^{1 \times d} \) is a learnable weight matrix.

For each time step \( n \), the input \( \mathbf{x}^{n}_{\text{tokenEmb}} \in \mathbb{R}^{C_{\text{Res}} \times d} \) undergoes a FFT along the channel dimension:
\begin{equation}
\mathbf{x}(f)^{n}_{\text{channel}} = \frac{1}{\sqrt{N_c}} \sum_{c=0}^{N_c-1} \mathbf{x}^{l}_{\text{tokenEmb}}(c) e^{-j \frac{2\pi}{N_c}cf}
\end{equation}
where \( f \) is the frequency index, \( c \) is the channel index, and \( N_c \) is the total number of channels. Due to the conjugate symmetry of real-valued signals, only the first half of the spectrum is retained. The resulting representation \( \mathbf{x}(f)^{l}_{\text{channel}} \in \mathbb{R}^{(C_{\text{Res}}/2) \times d} \) captures the frequency-domain features.

To model inter-channel dependencies, we apply MLPs that separately process the real and imaginary parts:
\begin{equation}
\begin{aligned}
\mathbf{y}^0 &= \mathbf{x}(f)^{n}_{\text{channel}}, n = 0 \\
\mathbf{y}^{n} &= \sigma\left(\text{Re}(\mathbf{y}^{n-1}) \mathbf{W}_r^n - \text{Im}(\mathbf{y}^{n-1}) \mathbf{W}_i^n + \mathbf{B}_r^n\right) \\
&\quad + j \sigma\left(\text{Re}(\mathbf{y}^{n-1}) \mathbf{W}_i^n + \text{Im}(\mathbf{y}^{n-1}) \mathbf{W}_r^n + \mathbf{B}_i^n\right) ,n > 0
\end{aligned}
\end{equation}
Here, \( \mathbf{W}^n = \mathbf{W}_r^n + j\mathbf{W}_i^n \in \mathbb{R}^{d \times d} \) and \( \mathbf{B}^n = \mathbf{B}_r^n + j\mathbf{B}_i^n \in \mathbb{R}^d \) denote the complex-valued weight matrices and biases for the \( n \)-th layer, and \( \sigma(\cdot) \) is the activation function.

An inverse Fast Fourier Transform (iFFT) is then applied to reconstruct the time-domain features:
\begin{equation}
\mathbf{z}^n(c) = \frac{1}{\sqrt{N_c}} \sum_{f=0}^{N_c-1} \mathbf{y}^n(f) e^{j \frac{2\pi}{N_c}cf}
\end{equation}
where \( \mathbf{z}^n(c) \in \mathbb{R}^{C_{\text{Res}} \times d} \) represents the reconstructed features in the time domain. Aggregating the reconstructed features across all time steps results in \( \mathbf{Z}(c) \in \mathbb{R}^{C_{\text{Res}} \times L_{\text{Res}} \times d} \).

To further capture inter-temporal dependencies, a frequency transformation is applied along the temporal dimension for each channel \( c \), using the input \( \mathbf{z}^c \in \mathbb{R}^{L_{\text{Res}} \times d} \). This yields the frequency-domain representation \( \mathbf{X}_{\text{fre}} \in \mathbb{R}^{C_{\text{Res}} \times L_{\text{Res}} \times d} \).

\begin{table*}[!t]
    \centering
    \captionsetup{aboveskip=2pt,font=small}
    \scriptsize
    \renewcommand{\arraystretch}{1}
    \caption{
        Overall Trajectory Prediction Accuracy and Ablation Study. 
        All metrics are in meters. Best results are in \textbf{bold}.
    }
    \renewcommand\cellalign{cc}
    \begin{tabular}{@{}ccccccccc|c|ccc@{}}
    \toprule
        \multirow{2}{*}{\textbf{Datasets}} & 
        \multirow{2}{*}{\textbf{Metric}} & 
        \multirow{2}{*}{\textbf{\makecell{RoNIN\\ResNet}}} &
        \multirow{2}{*}{\textbf{\makecell{RoNIN\\LSTM}}} &
        \multirow{2}{*}{\textbf{\makecell{RoNIN\\TCN}}} &
        \multirow{2}{*}{\textbf{RNIN}} &
        \multirow{2}{*}{\textbf{IMUNet}} &
        \multirow{2}{*}{\textbf{TLIO}} &
        \multirow{2}{*}{\textbf{FTIN}} &
        \multirow{2}{*}{\textbf{\makecell{FTIN vs\\RoNIN ResNet}}} &
        \multicolumn{3}{c}{\textbf{Ablation Study}} \\ 
        \cmidrule(l){11-13}
        & & & & & & & & & & \textbf{+FDL} & \textbf{+TDL} & \textbf{+FDL\&TDL} \\
        \midrule
        \multirow{2}{*}{RoNIN} & ATE & 6.799 & 6.136 & 5.500 & 5.686 & 5.550 & 4.417 & \textbf{3.875} & 43.0\% & 5.877 & 4.690 & \textbf{3.875} \\
                               & RTE & 3.523 & 3.595 & 3.563 & 3.449 & 3.387 & 3.507 & \textbf{3.061} & 13.1\% & 3.323 & 3.223 & \textbf{3.061} \\
        \multirow{2}{*}{RIDI}  & ATE & 2.608 & 3.294 & 2.743 & 1.851 & 2.645 & 2.641 & \textbf{1.727} & 33.8\% & 1.962 & 2.367 & \textbf{1.727} \\ 
                               & RTE & 2.881 & 3.136 & 2.476 & 2.040 & 2.788 & 3.105 & \textbf{1.981} & 31.2\% & 2.194 & 2.560 & \textbf{1.981} \\
        \multirow{2}{*}{RNIN}  & ATE & 1.182 & 1.893 & 1.561 & 1.603 & 1.417 & 1.493 & \textbf{1.096} & 7.3\%  & 1.193 & 1.097 & \textbf{1.096} \\ 
                               & RTE & 2.036 & 2.732 & 2.108 & 1.842 & 1.872 & 1.728 & \textbf{1.114} & 45.3\% & 1.199 & 1.123 & \textbf{1.114} \\
        \multirow{2}{*}{TLIO}  & ATE & 1.445 & 2.087 & 4.682 & 1.611 & \textbf{1.432} & 1.779 & 1.479 & -2.4\% & \textbf{1.383} & 1.448 & 1.479 \\ 
                               & RTE & 3.758 & 5.990 & 14.524 & 4.637 & \textbf{3.474} & 5.127 & 3.552 & 5.5\%  & 3.595 & 3.707 & \textbf{3.552} \\ 
        \multirow{2}{*}{OxIOD} & ATE & 3.557 & 4.277 & 26.442 & 4.343 & 9.822 & 5.071 & \textbf{3.446} & 3.1\%  & 3.518 & 3.452 & \textbf{3.446} \\ 
                               & RTE & 1.324 & 1.683 & 5.628 & 1.340 & 2.307 & 1.445 & \textbf{1.165} & 12.0\% & 1.308 & \textbf{1.165} & \textbf{1.165} \\ 
        \multirow{2}{*}{IMUNet}& ATE & 18.040 & 10.182 & 38.272 & 16.552 & 11.847 & 17.190 & \textbf{10.171} & 43.6\% & 14.610 & 17.557 & \textbf{10.171} \\ 
                               & RTE & 9.251 & 7.382 & 29.620 & 8.566 & 7.792 & 9.312 & \textbf{7.289} & 21.2\% & 8.598 & 7.809 & \textbf{7.289} \\ 
        \bottomrule
    \end{tabular}
    \label{results_combined}
    \vspace{-15pt}
\end{table*}

Next, the temporal and embedding dimensions are flattened to form \( \mathbf{X}_{\text{fre}} \in \mathbb{R}^{C_{\text{Res}} \times (L_{\text{Res}} \times d)} \). A fully connected layer followed by a nonlinear activation function is applied to enhance representational capacity. The final output of the FDL module is denoted as \( \mathbf{X}_{\text{fre}} \in \mathbb{R}^{C_{\text{fre}} \times L_{\text{fre}}} \), where \( C_{\text{fre}} \) and \( L_{\text{fre}} \) correspond to the output channel and temporal dimensions, respectively.

Through the coordinated operation of these components, the model emphasizes the most informative frequency components, reduces redundancy, and effectively captures long-range dependencies along both channel and temporal dimensions, thereby establishing a robust foundation for downstream localization tasks.

\begin{figure}[t] 
\centering
\captionsetup{aboveskip=2pt,font=small} 
\includegraphics[width=0.5\textwidth]{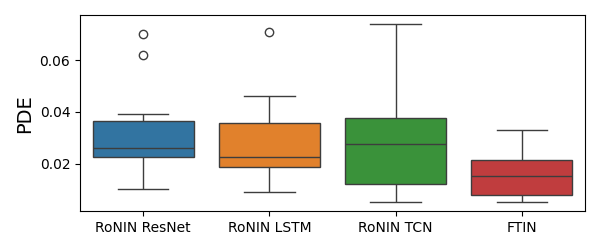}
\caption{PDE comparison between FTIN and three baseline models from RoNIN on the RoNIN dataset.}
\label{RoNIN}
\vspace{-20pt} 
\end{figure}

\subsection{Time-Domain Learning}
While FDL provides significant advantages, it does not eliminate the necessity of time-domain learning (TDL). In fact, the combination of frequency- and time-domain processing is a well-established strategy for capturing global context awareness in sequential signals. Therefore, the sLSTM module is introduced following the FDL stage. This module enhances global context modeling by optimizing its internal memory cell and adopting a scalar update mechanism. Such a design improves the efficiency of temporal sequence modeling, supports cross-domain information fusion, and captures temporal dependencies in IMU signals. In doing so, it enables the integration of complementary information from both the frequency and time domains.
The output of the time-domain learning stage is defined as:
\begin{equation}
X_{\text{temp}} = \text{sLSTM}(X_{\text{fre}}) \in \mathbb{R}^{C_{\text{temp}} \times L_{\text{temp}}}
\end{equation}
where $\text{sLSTM}(\cdot)$ denotes the sequence modeling operation performed by the sLSTM module.

\section{Experiments and Analysis}
\subsection{Experimental Settings}
\textbf{Datasets.} We utilize six commonly used open-source inertial datasets, including RIDI\cite{RIDI}, RoNIN\cite{RoNIN}, OxIOD\cite{Oxiod}, RNIN\cite{RNIN-VIO}, TLIO\cite{TLIO}, and IMUNet\cite{IMUNet}. Prior to training, each dataset is randomly shuffled and split into training, validation, and testing sets in an 8:1:1 ratio to ensure fair and consistent evaluation. 

\textbf{Metric.}
We adopt three widely used evaluation metrics to assess localization performance: Absolute Trajectory Error (ATE)\cite{IDOL}, Relative Trajectory Error (RTE)\cite{IDOL}, and Position Drift Error (PDE)\cite{CTIN}. ATE and RTE evaluate the model’s accuracy in predicting global and local trajectories, respectively Lower values of these metrics indicate better localization performance. PDE quantifies the drift error at the endpoint of the reconstructed trajectory.

\textbf{Baselines.} Compared to traditional Newtonian-based methods, machine learning approaches have demonstrated significant performance advantages\cite{SurveyofIndoorInertial,surveyILS}. Therefore, we primarily compare several representative learning-based models, including the ResNet\cite{ResNet}, LSTM\cite{LSTM}, and TCN\cite{TCN} architectures introduced in RoNIN~\cite{RoNIN}. We also include IMUNet\cite{IMUNet}, as well as the networks employed in TLIO\cite{TLIO} and RNIN\cite{RNIN-VIO}, as comparative baselines.

\begin{figure*}[!t] 
\centering
\captionsetup{aboveskip=2pt,font=small} 
\includegraphics[width=1.0\textwidth]{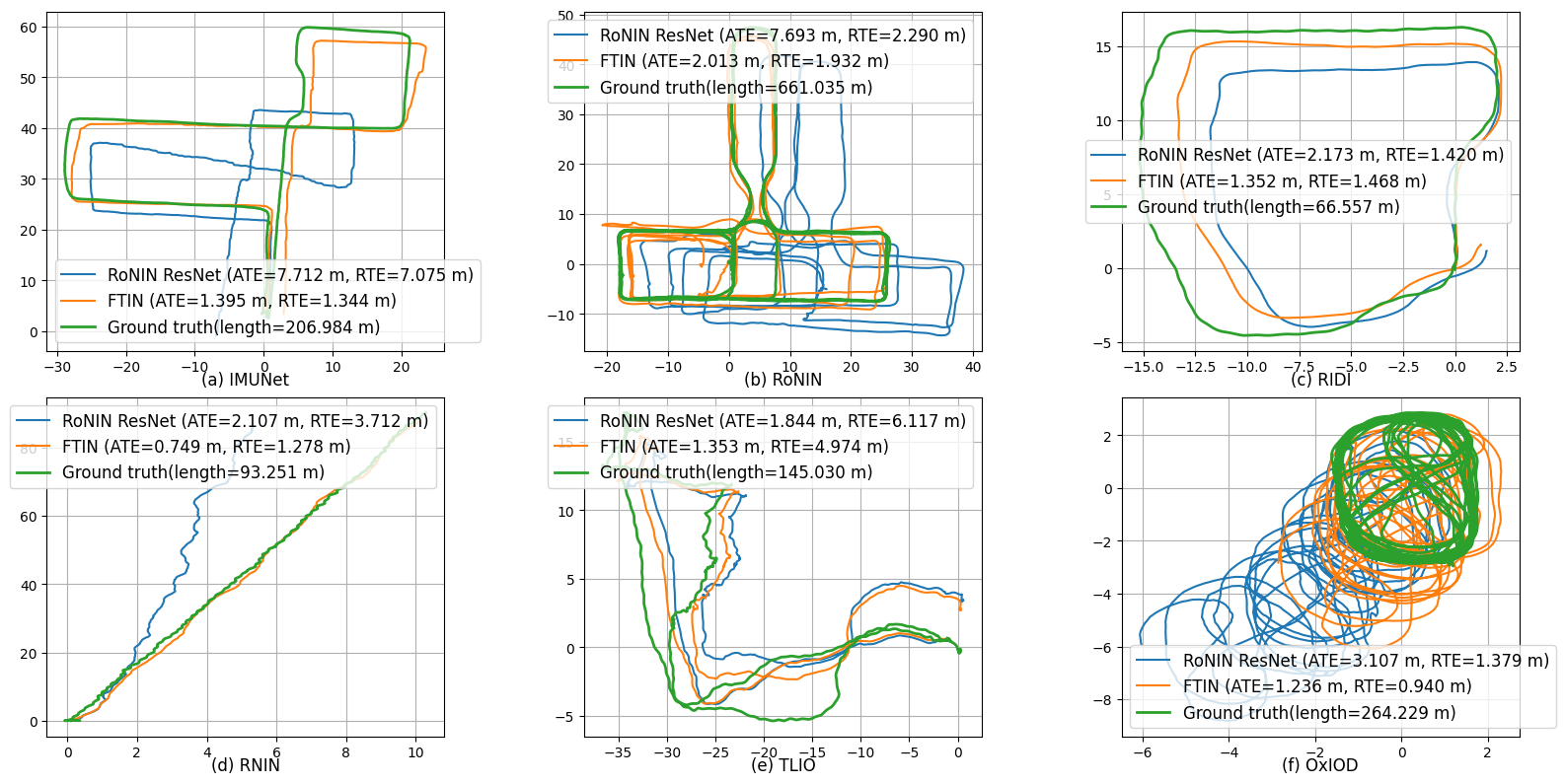}
\caption{Visual comparison of the FTIN and RoNIN ResNet on various datasets.}
\label{visual trajectory}
\vspace{-20pt} 
\end{figure*}

\textbf{Training Details.} All training and testing were conducted on an NVIDIA RTX A40 GPU (48GB). The batch size was set to 512, with a maximum of 100 training epochs. The initial learning rate was set to $10^{-4}$, and the Adam optimizer was employed for model training. Early stopping was triggered when the learning rate decayed below $10^{-6}$ to mitigate the risk of overfitting.

\subsection{Comparisons with the Baselines}
\textbf{Overall Performance.}
The left portion of Table~\ref{results_combined} reports the ATE and RTE for various models across the test datasets. The results indicate that FTIN consistently outperforms other network architectures in both ATE and RTE across most scenarios. On the RoNIN dataset, FTIN achieves reductions of 43.0\% in ATE and 13.1\% in RTE compared to RoNIN ResNet. Notably, the RoNIN-based models (ResNet, LSTM, and TCN), along with RNIN, IMUNet, and TLIO, rely solely on time-domain representations. This lack of frequency-domain modeling results in an information bottleneck when processing high-frequency IMU signals, ultimately limiting localization accuracy.

Furthermore, Fig.~\ref{RoNIN} presents the PDE for FTIN and the baseline methods. FTIN achieves the lowest PDE, demonstrating its superior capability in approximating trajectory endpoints. Overall, FTIN enables cross-domain information fusion and effectively overcomes the information bottleneck, delivering substantial improvements in localization accuracy across multiple benchmark datasets. These findings further validate its effectiveness for IO.

\textbf{Visualization.}
Fig.~\ref{visual trajectory} provides a visual comparison between FTIN and the baseline models using representative trajectory samples from each test set. In Fig.~\ref{visual trajectory} (c) and (d), the accumulation of heading errors in the baseline models is clearly evident. In contrast, FTIN exhibits better trajectory-fitting performance in short-range motion scenarios. In Fig.~\ref{visual trajectory} (b) and (f), substantial rotational motion causes the baseline models to gradually deviate from the ground-truth trajectory. While FTIN maintains robust tracking in Fig.~\ref{visual trajectory} (b), its performance in Fig.~\ref{visual trajectory} (f) still leaves room for improvement. In other cases, FTIN generally produces accurate displacement estimation.
These visual results further support the effectiveness of combining time-domain and frequency-domain learning strategies to enhance the accuracy of IO.

\subsection{Ablation Study}
To further validate the effectiveness of the proposed method, we conduct ablation studies on the FDL and TDL modules. The right side of Table~\ref{results_combined} presents the ablation results on the open-source datasets used in our experiments. We progressively incorporate the FDL and TDL modules into the ResNet-1D backbone. The results indicate that both modules independently contribute to improving trajectory estimation accuracy, as reflected by reductions in ATE and RTE. Moreover, when FDL and TDL are combined, they yield lower localization errors in the vast majority of cases, confirming both the effectiveness and complementarity of the two components.

\section{Conclusion}
This paper investigates the IO problem from a frequency-domain perspective. By fusing information from both the frequency and time domains, the proposed method alleviates information bottlenecks and substantially improves trajectory estimation accuracy. The resulting dual-domain approach offers a novel perspective for advancing IO research. Nevertheless, the extent to which IMU signals vary across platforms remains an open question and warrants further investigation.

\section{Acknowledgements}
This work is supported by Science and Technology Major Program of Fujian Province (No. 2022HZ026007) and partly supported by the Science and Technology Planning Project of Fujian province (2022I0001).

\vfill\pagebreak

\bibliographystyle{IEEEbib}
\bibliography{refs}

\end{document}